\documentclass[conference]{IEEEtran}
\IEEEoverridecommandlockouts
\usepackage{amsmath,amssymb,amsfonts}
\usepackage{algorithmic}
\usepackage{graphicx}
\usepackage{textcomp}
\usepackage{xcolor}

\usepackage{biblatex} 
\def\BibTeX{{\rm B\kern-.05em{\sc i\kern-.025em b}\kern-.08em
    T\kern-.1667em\lower.7ex\hbox{E}\kern-.125emX}}

\usepackage{tikz}
\usetikzlibrary {backgrounds, fit, calc} 
\usepackage{qcircuit}
\DeclareUnicodeCharacter{0301}{O}
\usepackage{balance}
\balance

\begin{document}

\title{QNet: A Quantum-native Sequence Encoder Architecture}

\author{\IEEEauthorblockN{Wei Day}
\IEEEauthorblockA{\textit{Department of Computer Science} \\
\textit{National Central University}\\
Taoyuan City, Taiwan \\
academic@davidday.tw}
\and
\IEEEauthorblockN{Hao-Sheng Chen}
\IEEEauthorblockA{\textit{Department of Computer Science} \\
\textit{National Central University}\\
Taoyuan City, Taiwan \\
cliffxzx@gmail.com}
\and
\IEEEauthorblockN{Min-Te Sun}
\IEEEauthorblockA{\textit{Department of Computer Science} \\
\textit{National Central University}\\
Taoyuan City, Taiwan \\
msun@csie.ncu.edu.tw}
}

\maketitle

\begin{abstract}
This work proposes QNet, a novel sequence encoder model that entirely inferences on the quantum computer using a minimum number of qubits. Let $n$ and $d$ represent the length of the sequence and the embedding size, respectively. The dot-product attention mechanism requires a time complexity of $O(n^2 \cdot d)$, while QNet has merely $O(n+d)$ quantum circuit depth. In addition, we introduce ResQNet, a quantum-classical hybrid model composed of several QNet blocks linked by residual connections, as an isomorph Transformer Encoder. We evaluated our work on various natural language processing tasks, including text classification, rating score prediction, and named entity recognition. Our models exhibit compelling performance over classical state-of-the-art models with a thousand times fewer parameters. In summary, this work investigates the advantage of machine learning on near-term quantum computers in sequential data by experimenting with natural language processing tasks.
\end{abstract}


\section{Introduction}

Quantum computer hardware technologies are maturing, making quantum algorithms a possible solution to reduce the training cost of neural networks. When this paper was written, a quantum computer with 127 qubits was revealed~\cite{chow2021ibm}. Multiple reports~\cite{huang2022quantum, schuld2022quantum} show that Quantum Machine Learning (QML) outperforms classical machine learning in certain situations, but the actual capability of QML remains to be discovered. Most QML models are built on top of Variational Quantum Circuits (VQC), the quantum algorithms that depend on free parameters. Based on VQC, multiple quantum Feedforward neural networks capable of universal quantum computation have been proposed~\cite{schuld2014quest, wan2017quantum, farhi2018classification, mitarai2018quantum, beer2020training} to exploit the potential of quantum computers in machine learning and attempt to train networks more effectively.

It is also worth mentioning that utilizing quantum computers for machine learning is a challenging task. Other than the gradient vanishing or exploding problem that already exists in classical computers, quantum computers will incur decoherence after a long period of execution in the noisy intermediate-scale quantum (NISQ) era~\cite{RevModPhys.76.1267}. In addition, the resource of quantum computing is limited as quantum computers are still under development. Quantum machine learning researchers must propose models that can both be verified on NISQ devices and are extendable for large-scale quantum computers in the future.

On the other hand, machine learning is another fast-growing domain that is good at recognizing patterns and solving many real-life problems. A sequence-to-sequence deep learning model, Transformer~\cite{NIPS2017_3f5ee243} architecture, has achieved rapid and widespread dominance in various tasks, especially the models that use Transformer Encoder as their main component, such as the BERT~\cite{devlin2018bert} model and Vision Transformer~\cite{dosovitskiy2021an}. Each block in the Transformer Encoder comprises a Multi-head self-attention and a Feedforward layer. The former is an inductive bias that connects each token in the input through a relevance-weighted basis of every other token, and the latter is a parameterized linear transformation on each token.

However, one crucial shortcoming of Transformer is the time complexity of Multi-head self-attention. It performs a matrix multiplication operation that needs the time complexity of $O(n^2 \cdot d)$ to relate the tokens, where $n$ and $d$ represent sequence length and embedding dimensions, respectively. This makes the time cost to input the long sequence to the model impractical. Multiple works~\cite{arxiv.2006.04768, tolstikhin2021mlp, melas2021you, DBLP:journals/corr/abs-2105-03824, NEURIPS2021_4cc05b35} have attempted to enhance Transformer Encoder to improve either performance or accuracy, and in our observations, accuracy is often sacrificed for speed.

In this work, we propose QNet, a sequence encoder block inspired by the Transformer encoder that entirely inferences on quantum computers. A sequence encoder is a model that reads the input sequence and summarizes the information into a context vector representing a certain meaning of the input, which is usually the prediction target of a task. In addition to QNet, ResQNet, the model composed of residual connected QNet blocks, is proposed. ResQNet has shown state-of-the-art performance while reducing the complexity of the model on various natural language processing tasks.

In the experiment sections, four datasets are used for evaluation. On the ColBERT sentence emotion classification dataset, QNet and ResQNet resulted in an accuracy of 89.84\% and 91.17\% with dozens of parameters, while other models have approximate results but with parameters of higher magnitudes. On the Stackoverflow Question classification dataset, QNet and ResQNet achieve higher accuracy than other models. For the MSRA named-entity recognition task, experimented models are challenged to assign labels to every input token in a heavily imbalanced distribution. However, QNet and ResQNet both show compelling results on the MSRA dataset, proving the effectiveness of such architecture design.
\section{Preliminary}

\subsection{Attention In Sequential Task}

With the development of deep learning, Natural Language Processing (NLP) technology has become progressively mature. NLP is a technology that enables computers to analyze human language, which has a sequential context relationship. Series data is traditionally processed with Recurrent Neural Networks (RNN) that have difficulty performing parallel operations and easily face gradient vanishing issues. Thus, Google proposed a network architecture that does not use RNN and CNN but only the self-attention mechanism - Transformer~\cite{NIPS2017_3f5ee243}, in which each block is composed of Multi-head attention and a Feedforward layer. In recent years, Transformer has achieved outstanding results in tasks such as sentiment analysis~\cite{naseem2020transformer, wang2020transmodality}, machine translation~\cite{vaswani2018tensor2tensor}, speech recognition~\cite{dong2018speech, gulati2020conformer}, and dialogue robots~\cite{zandie2020emptransfo, suglia2021embodied}.

The Transformer architecture has inspired multiple large-scale NLP models. For instance, Bidirectional Encoder Representations from Transformers (BERT)~\cite{devlin2018bert} is developed by Google for NLP pre-training, mainly composed of a stack of Transformer encoders and a classification layer. It is designed to help computers understand the meaning of ambiguous language in the text by using surrounding text to establish context.

Equation~\ref{equ:attention} is the attention mechanism that flexibly outputs the most relevant parts of the input sequence, $Q$ and $K$ are vectors used to compute the relation weight of every two tokens, and $V$ is a vector used to add the weight to retrieve the relation information. These vectors usually come from the same source but are weighted linear transformed to introduce differences to each.

\begin{equation} \label{equ:attention}
    Attention(Q, K, V) = softmax(\frac{QK^T}{\sqrt{d_k}})V
\end{equation}

Recent works have shown the potential mathematics transformations that can replace Multi-head self-attention and achieve approximately the same performance. In fact, the approach of replacing self-attention with a transpose operation and a Feedforward layer has proved to be able to achieve convincing results.~\cite{tolstikhin2021mlp}

The work in~\cite{DBLP:journals/corr/abs-2105-02723} starts the question about the primitive property of the attention mechanism. The work in~\cite{DBLP:journals/corr/abs-2105-03824} shows that standard un-parameterized discrete Fourier transform can speed up Transformer encoder architectures, with limited accuracy costs, by replacing the self-attention sublayer with simple linear transformations that \emph{mix} input tokens.

\subsection{Quantum Machine Learning}

Quantum machine learning is a category of machine learning that makes use of quantum computers, that is, utilizing the properties of qubits and quantum gates. Instead of building machine learning models that are purely supported on quantum computers, models designed in the quantum-classical hybrid structure are more common for NISQ devices. One of the reasons is that the input data are permanently stored on classical computers because the decoherence of quantum computers will corrupt the data. In quantum-classical hybrid models, the quantum computer is either used to provide the quantum data by extracting input features or employing a quantum machine learning algorithm as the classifier.

In this work, our models are used as representation encoders that pipe the output to a classical fully-connected layer. However, QNet is not restricted to being a backbone of classical classifiers since it can be directly connected to a quantum neural network to make a pure quantum model.

\subsubsection{Quantum Fourier Transform}
Quantum Fourier transform~\cite{nielsen2002quantum} is the quantum implementation of the discrete Fourier transform over the amplitudes of a quantum state that converts the amplitudes from the time domain to the frequency domain.

Similar to discrete Fourier transform, Quantum Fourier transform maps the quantum state from $|X\rangle = \sum_0^{N-1} x_j|j\rangle$ to $|Y\rangle = \sum_0^{N-1} y_k |k\rangle$ according to Equation~\ref{equ:qft}, where $\omega^{jk}_N = e^{2\pi i \frac{jk}{N}}$.

\begin{equation} \label{equ:qft}
y_k = \frac{1}{\sqrt{N}} \sum_0^{N-1} x_j \omega_N^{jk}
\end{equation}

The unitary matrix can also express Quantum Fourier transform as Equation~\ref{equ:qft_u} with $\omega$ defined the same as above.

\begin{equation} \label{equ:qft_u}
U_{QFT} = \frac{1}{\sqrt{N}} \sum_{j=0}^{N-1}  \sum_{k=0}^{N-1} \omega_{N}^{jk} |k\rangle \langle j|
\end{equation}

This work uses the Quantum Fourier Transform in the "Mixture Learning" layer to blend multiple individual qubits representing the same embedding dimension with multiple tokens to create an information-mixed intermediate state.

\subsubsection{Variational Quantum Eigensolver}

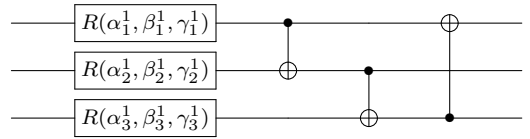
\begin{figure}[htp]
  \centerline {
    \footnotesize
    \Qcircuit @C=3em @R=.5em {
        & \gate{R(\alpha^1_1, \beta^1_1, \gamma^1_1)} & \ctrl{1} & \qw & \targ & \qw \\
        & \gate{R(\alpha^1_2, \beta^1_2, \gamma^1_2)} & \targ & \ctrl{1} & \qw & \qw \\
        & \gate{R(\alpha^1_3, \beta^1_3, \gamma^1_3)} & \qw & \targ & \ctrl{-2} & \qw
    }
   }
  \caption{Classic single-Layer VQE with 3 quantum wires.}
  \label{fig:vqe}
\end{figure}

The Variational Quantum Eigensolver (VQE)~\cite{peruzzo2014variational} is a form of the quantum circuit with configurable parameters iteratively tuned by a classical computer. It can provide solutions in regimes that lie beyond the research of conventional algorithms. VQE finds an optimal transformation in a set of unitary gates that depends on the design of the quantum circuit. Fig.~\ref{fig:vqe} presents a sample VQE circuit where $\alpha, \beta, \gamma$ are sets of learnable parameters to rotate qubits in 3 axes. Entanglement gates usually follow rotation gates to let qubits interact with each other to exchange information.

The VQE approach has been shown to be flexible in circuit depth and insensitive to the presence of noises~\cite{zeng2021simulating}. Therefore, while there is still a lack of quantum error correction and fault-tolerant quantum computation in the NISQ era, quantum machine learning methods driven by variational quantum circuits can circumvent the complex quantum flaws in the current quantum devices.

\subsection{Word Embedding Sparsity}

Multiple works~\cite{10.5555/3060832.3061029}~\cite{Panahi2020word2ket:} have shown evidence that the information contained in the word embedding is sparse and can be factored with many fewer elements. The word embedding structure inspired by quantum entanglement, word2ket~\cite{Panahi2020word2ket:} is one of the most fantastic works. The experimental results show that after a 34,000-fold reduction in trainable parameters, it can still match the scores of the original unreduced configurations.

These works indicate the potential of using fewer embedding dimensions to achieve state-of-the-art performance, especially by utilizing quantum entanglement. The model interpretability can also be profitable by examining fewer parameters.

\section{Related Work}

\subsection{Quantum Recurrent Neural Network}

\begin{figure}[htbp]
\centerline{\includegraphics[width=9cm]{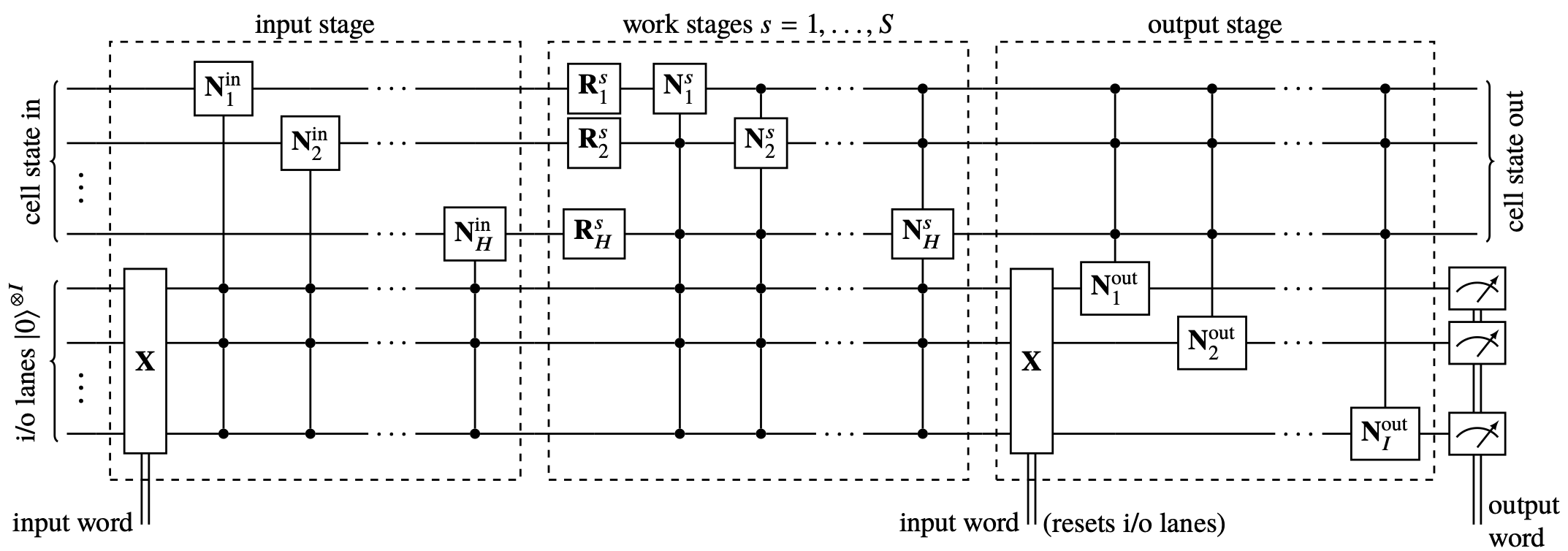}}
\caption{The cell of Quantum Recurrent Neural Network, the \textbf{N} gates are the controlled quantum neurons.}
\label{qrnn-cell}
\end{figure}

Prior to this work, the quantum recurrent neural network (QRNN) is proposed as a pure quantum version implementation of recurrent neural networks.~\cite{NEURIPS2020_0ec96be3} Fig.~\ref{qrnn-cell} is the QRNN cell where qubits are divided into hidden state wires, which hold the hidden state of former cells and pass to latter units, and input state wires, the temporary wires for the entanglement gates to integrate input to the hidden state.

The work of QRNN shows the benefits of pure quantum neural network architectures. It can be used as a generative model based on the nature of quantum gates that is unitary and can analyze what extent the unitary nature of the network counteracts the vanishing gradient problem.

However, in the NISQ era, the entangled qubits to decoherence could be easily caused by the noises in deep circuits, so it might be unfeasible to employ QRNN in near-term quantum devices since the circuit depth of the QRNN is $O(n \times d)$ in which $n$ is the length of the sequence and $d$ is the dimension of the hidden state.

\subsection{Quantum Long Short-Term Memory}

\begin{figure}[htbp]
\centerline{\includegraphics[width=7cm]{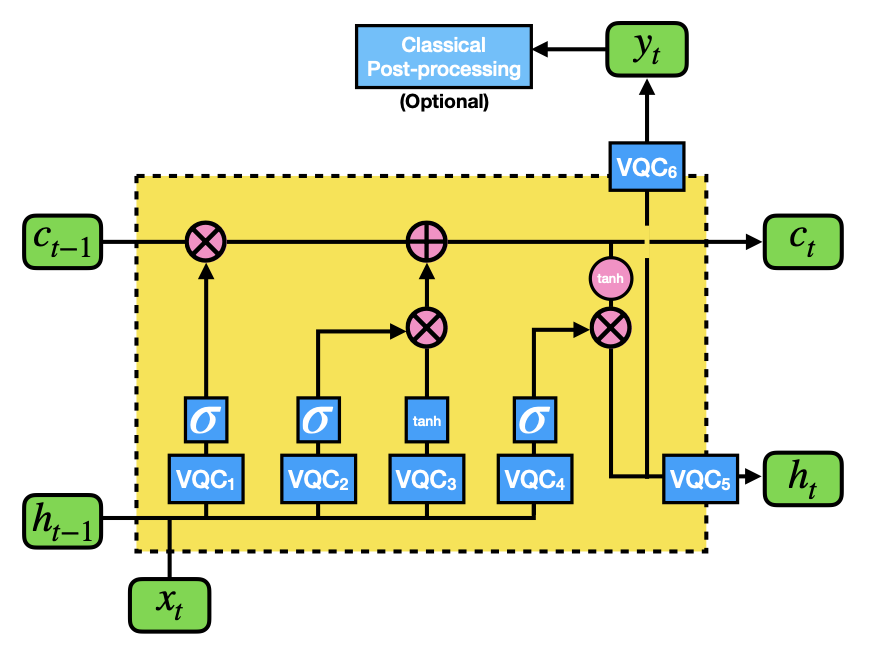}}
\caption{The cell of Quantum Long-Short-Term Memory.}
\label{qlstm-cell}
\end{figure}

In the proposed Quantum Long Short-Term Memory (QLSTM), hybrid quantum computing is being used to enhance the data before or after gates in the Long Short-Term Memory cell by conducting transformation with VQE. The cell of QLSTM is shown in Fig.~\ref{qlstm-cell}.

The work of QLSTM shows that, in some cases, the predictions of QLSTM can fit better and converge much faster than LSTM. But the measurement operation on quantum computers is time expensive, and each VQE in QLSTM requires measurement to continue to compute other parts of the cell in the classical computer. Since the model performance improvement is slight, it is not beneficial to employ such architecture in a hybrid system of quantum computers and classical computers that would have bottlenecks in transferring data between them.

\subsection{Quantum Enhanced Transformer}

Transformer architecture revolutionized the analysis of sequential data. It is intuitive to attempt to integrate such a state-of-the-art model with quantum computers to seek improvement. The work~\cite{DBLP:journals/corr/abs-2110-06510} that proposed Q-Transformer is inspired by QLSTM and is also a hybrid model in which VQE is used to enhance the input of the Attention.

The inputs of the self-attention are often transformed with weight as Equ.~\ref{equ:weighted-attention}. The concept of Q-Transformer is to replace the linear transformation $W_Q$, $W_K$, and $W_V$ with three VQE, respectively.

\begin{equation} \label{equ:weighted-attention}
    Attention(QW_Q, KW_K, VW_V)
\end{equation}

The Q-Transfomer was experimented on the IMDB dataset and took about 100 hours to train a single epoch. However, in the end, the author did not present the result of training Q-Transformer.

\section{Method}

\subsection{Problem Formulation}

This paper aims to design a quantum machine learning model with a similar architecture to BERT, a sequence-to-sequence model, and to explore how much improvement it can achieve. Using the sequence-to-sequence pre-trained model as a backbone can also adapt the model to other tasks, such as text classification or image segmentation.

The methods of this paper are proposed under the following assumptions.

\begin{itemize}
  \item The NISQ device can compute any general gate in $O(1)$ time complexity. General gates include two standard entanglement gates (CNOT and SWAP) and arbitrary rotation gates.
  \item There is a finite number of qubits on the NISQ device.
  \item The model inputs can be chopped and padded into the same length.
\end{itemize}

\subsection{QNet}

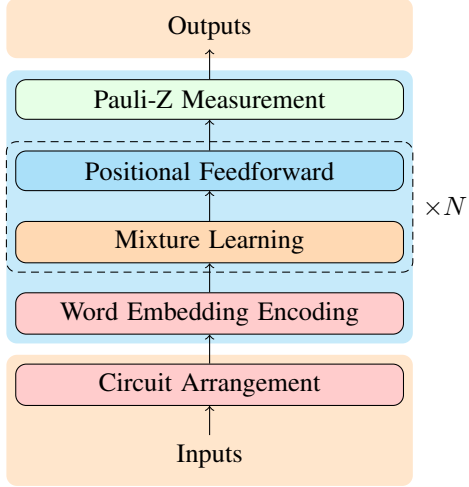
\begin{figure}[htp!]
  \centering
    \tikzstyle{layer} = [draw, text centered, text width=14em, minimum height=1.5em, rounded corners]
    \tikzstyle{ffblock} = [layer, fill=cyan!30]
    \tikzstyle{qftblock} = [layer, fill=orange!30]
    \tikzstyle{measure} = [layer, fill=green!10]
    \tikzstyle{encode} = [layer, fill=red!20]
    \tikzstyle{params} = [layer, fill=red!20]
    \begin{tikzpicture}[node distance=2.7em]
        \node (output) [text centered, text width=14em] {Outputs};
        \node (measure) [measure] [below of=output] {Pauli-Z Measurement};
        \node (ffblock) [ffblock] [below of=measure] {Positional Feedforward};
        \node (qft) [qftblock] [below of=ffblock] {Mixture Learning};
        \node (encode) [encode] [below of=qft] {Word Embedding Encoding};
        \node (parameters) [params] [below of=encode] {Circuit Arrangement};
        \node (input) [text centered, text width=14em] [below of=parameters] {Inputs};
        
        \begin{scope}[on background layer]
            \node [fill=cyan!20, rounded corners, fit={(encode) (qft) (ffblock) (measure) }] {};
            \node [fill=orange!20, rounded corners, fit={(input) (parameters)}] {};
            \node [fill=orange!20, rounded corners, fit={(output)}] {};
            \node [draw, densely dashed, rounded corners, fit={(qft) (ffblock)}, label={right:$\times N$}] {};
            
            \draw[->] (input) -- (parameters);
            \draw[->] (parameters) -- (encode);
            \draw[->] (encode) -- (qft);
            \draw[->] (qft) -- (ffblock);
            \draw[->] (ffblock) -- (measure);
            \draw[->] (measure) -- (output);
        \end{scope}
        
    \end{tikzpicture}
  \caption{The QNet Encoder Block Architecture}
  \label{fig:qnet-encoder}
\end{figure}

The QNet encoder architecture is shown in Fig.~\ref{fig:qnet-encoder}. The blocks within the cyan box are the components that execute on quantum computers and the blocks within the orange box are the components that perform on classical computers. First, the "Circuit Arrangement" layer prepares the parameters the quantum circuit requires and assigns gates to both input and parameters. Second, the "Word Embedding Encoding" layer encodes the inputs into qubits. Then, the "Mixture Learning" layer introduces the mechanism to learn the token relationships. Next, the "Positional Feedforward" layer performs a learnable transformation on each token. Finally, qubits are measured to return the quantum data to the classical computer. Note that the dot-surrounded area can be repeated multiple times to increase the parameters and depth of the model.

The derivation and circuit design are introduced in detail in the following sections.

\subsubsection{Word Embedding Encoding}

Without modifying embeddings to encode positional information like \emph{Positional Encoding} in Transformer, QNet directly utilizes another axis of the qubit to store the positional information. Word embeddings are transformed to the volume of a qubit, the rotation on Pauli-X. The position of the word is encoded as the phase of the qubit, the angle on Pauli-Z, so the positional information is the distributed angles in the range of $[0, \pi]$

\begin{equation} \label{equ:word_embedding}
x \rightarrow |\psi_x\rangle = \prod_{i=1}^n \prod_{j=1}^d {R^{i*d+j}_z( \frac{(i-1) \pi}{n} ) R^{i*d+j}_x(x^i_j)}\;|0^{\otimes n\times d}\rangle
\end{equation}

The transformation of the quantum state can be described by Equation~\ref{equ:word_embedding}. In the equation, $x$ is the input token array, the superscript denotes the index of the token, and the subscript denotes the index of the embedding. Here, $n$ is the length of the input sequence, and $d$ is the embedding dimension. 


\begin{figure}[htp!]
  \centerline {
    \footnotesize
    \Qcircuit @C=3em @R=.4em {
     & \lstick{|0\rangle} & \gate{R_x(x^1_1)} & \gate{R_z(0)} & \rstick{|\psi^1_1\rangle} \qw \\
     & \lstick{|0\rangle} & \gate{R_x(x^2_1)} & \gate{R_z(0)} & \rstick{|\psi^2_1\rangle} \qw \\
     & \lstick{|0\rangle} & \gate{R_x(x^1_2)} & \gate{R_z(\frac{\pi}{2})} & \rstick{|\psi^1_2\rangle} \qw \\
     & \lstick{|0\rangle} & \gate{R_x(x^2_2)} & \gate{R_z(\frac{\pi}{2})} & \rstick{|\psi^2_2\rangle} \qw
    }
  }
  \caption{The quantum circuit of Word Embedding Quantum Encoding for 2 words with 2 embeddings.}
  \label{fig:embedding}
\end{figure}
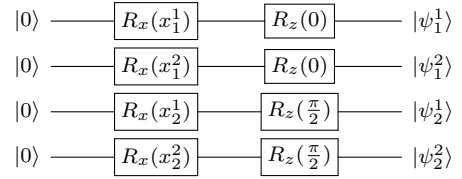

\subsubsection{Mixture Learning}

A self-attention mechanism is an attempt to implement the action of selectively concentrating on a few relevant things while ignoring others in deep neural networks. However, directly implementing dot-product attention in the Transformer Encoder is impractical to realize in the current scale of NISQ devices because, after the operation, the swap test will corrupt qubits and require duplicate qubits for further calculation, which will need at least an additional $O(n \times d)$ number of qubits.

Instead of replicating the original self-attention, this work uses an alternative solution that is based on the different waveforms generated by Quantum Fourier Transform. The unitary of Quantum Fourier Transform is presented in Equation~\ref{equ:u_qft}, where $n$ is the length of the input sentence and the $N$-th root of unitary is defined as $\omega^{jk}_N = e^{2\pi i \frac{jk}{N}}$.

\begin{equation} \label{equ:u_qft}
U_{QFT} = \frac{1}{\sqrt{n}} \sum_{j=0}^{n-1}  \sum_{k=0}^{n-1} \omega _{n}^{jk} |k\rangle \langle j|
\end{equation}

Take an input length of 4 as an example. The matrix of the $U_QFT$ under the condition of $n = 4$ is presented in Equation~\ref{equ:matrix_qft}. After applying the unitary, the quantum state would become a mixture where each original state is selected with the weight of a waveform amplitude.

\begin{equation} \label{equ:matrix_qft}
U_{QFT} |\psi\rangle =
\begin{bmatrix}
    x_1 + x_2 + x_3 + x_4 \\
    x_1 + x_2 \omega_n^1 + x_3 \omega_n^2 + x_4 \omega_n^3 \\
    x_1 + x_2 \omega_n^2 + x_3 \omega_n^4 + x_4 \omega_n^6\\
    x_1 + x_2 \omega_n^3 + x_3 \omega_n^6 + x_4 \omega_n^9
\end{bmatrix}
\end{equation}

To apply a learnable relation weight transformation, single qubit rotation, as shown in Equation~\ref{equ:single_qubit}, is applied to every qubit involved in the mixture learning with the same weight. The whole unitary can be written as Equation~\ref{equ:span_Uw} where $\otimes$ is the Kronecker product operator and $w$ are trainable parameters. The reason that the exact weight is being used is that we do not wish to modify the original quantum state too much. For instance, if there is no requirement to have information on relationships, the $U_W'$ should be an identical matrix $I$, so instead of learning $O(2^n)$ parameters, optimizing a constant number of parameters to $I$ is much easier.

\begin{equation} \label{equ:single_qubit}
U_W = \begin{bmatrix}
 w_1 & w_2 \\
 w_3 & w_4
\end{bmatrix}
\end{equation}

\begin{equation} \label{equ:span_Uw}
 U'_W = U_W \otimes U_W \otimes ... U_W \otimes U_W
\end{equation}

After the learnable transformation, the inverse Quantum Fourier Transform unitary, represented by $U_{QFT}^\dag$, is applied. If nothing is learned, the quantum state would be transformed back to the same before $U_{QFT}$, or else such transformation would let qubits carry additional information from other tokens. The overall mixture learning operation can be presented as Equation~\ref{equ:mixture_learning}.

\begin{equation} \label{equ:mixture_learning}
|\psi'\rangle = U_{mix} |\psi\rangle = U_{QFT}^\dag U'_W U_{QFT} |\psi\rangle
\end{equation}

In order to create a mixture state of different embedding dimensions, Mixture Learning is employed for every single embedding dimension. The formula is shown as Equation~\ref{equ:mixture_learning_ops} where $d$ is the embedding dimension.
 
\begin{equation} \label{equ:mixture_learning_ops}
|\psi'\rangle = U_{mix_1}\ldots U_{mix_d}|\psi\rangle
\end{equation}

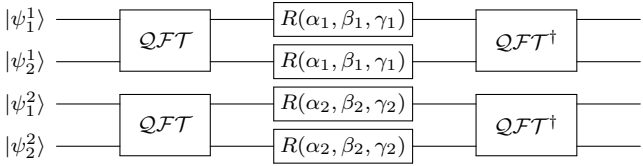
\begin{figure}[htp!]
  \centerline{
    \footnotesize
    \Qcircuit @C=3em @R=.4em {
     & \lstick{|\psi^1_1\rangle} & \multigate{1}{\ \mathcal{QFT}\ } & \gate{R(\alpha_1, \beta_1, \gamma_1)} & \multigate{1}{\ \mathcal{QFT^\dag}\ } & \qw \\
     & \lstick{|\psi^1_2\rangle} & \ghost{\ \mathcal{QFT}\ } & \gate{R(\alpha_1, \beta_1, \gamma_1)} & \ghost{\ \mathcal{QFT^\dag}\ } & \qw \\
     & \lstick{|\psi^2_1\rangle} & \multigate{1}{\ \mathcal{QFT}\ } & \gate{R(\alpha_2, \beta_2, \gamma_2)} & \multigate{1}{\ \mathcal{QFT^\dag}\ } & \qw \\
     & \lstick{|\psi^2_2\rangle} & \ghost{\ \mathcal{QFT}\ } & \gate{R(\alpha_2, \beta_2, \gamma_2)} & \ghost{\ \mathcal{QFT^\dag}\ } & \qw
    }
  }
  \caption{Circuit of Mixture Learning layer.}
  \label{fig:mixture_learning}
\end{figure}

Therefore, the Mixture Learning layer is made up of Quantum Fourier Transforms with parameterized transformation performed on all tokens for each qubit in the embedding. The trainable quantum circuit is shown in Fig.~\ref{fig:mixture_learning}, where $\alpha, \beta, \gamma$ are parameters that use rotation gates to operate $U_W$.

\subsubsection{Positional Feedforward}

\begin{figure}[htp!]
  \centerline{
    \footnotesize
    \Qcircuit @C=1em @R=.4em {
     & \lstick{|\psi^1_1\rangle} & \gate{R(\alpha_3,\beta_3,\gamma_3)} & \multigate{1}{\mathcal{G}} & \gate{R(\alpha_4,\beta_4,\gamma_4)} & \multigate{1}{\mathcal{G}} & \qw \\
    & \lstick{|\psi^2_1\rangle} & \gate{R(\alpha_5,\beta_5,\gamma_5)} & \ghost{\mathcal{G}} & \gate{R(\alpha_6,\beta_6,\gamma_6)} & \ghost{\mathcal{G}} & \qw \\
     & \lstick{|\psi^1_2\rangle} & \gate{R(\alpha_3,\beta_3,\gamma_3)} & \multigate{1}{\mathcal{G}} & \gate{R(\alpha_4,\beta_4,\gamma_4)} & \multigate{1}{\mathcal{G}} & \qw \\
    & \lstick{|\psi^2_2\rangle} & \gate{R(\alpha_5,\beta_5,\gamma_5)} & \ghost{\mathcal{G}} & \gate{R(\alpha_6,\beta_6,\gamma_6)} & \ghost{\mathcal{G}} & \qw
    }
  }
  \caption{QNet Feedforward with two wires.}
  \label{fig:feedforward}
\end{figure}
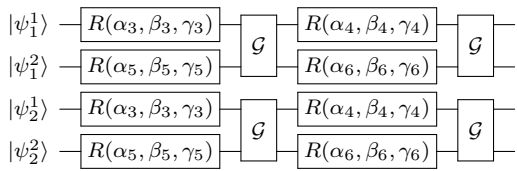

In this work, we propose a special feedforward model that aims to construct a viable machine-learning model with the existing NISQ-era technology. Unlike traditional quantum neural networks~\cite{beer2020training}, which are known to have a three-dimensional architecture with each neuron represented by a qubit and generating plenty of auxiliary qubits, the feedforward layer used in QNet abandons the intermediate qubits to simplify the Feedforward layer using the structure inspired by Grover's algorithm, and the circuit is shown in Fig.~\ref{fig:feedforward}.

In a typical transformer architecture, the Feedforward layer has two fully connected layers with one activation function in between. Here, we use a similar arrangement by putting two VQEs separated by two transformation functions $\mathbf {G}$ surrounding the latter VQE circuit. The process can be demonstrated as Equation~\ref{equ:Feedforward}.

\begin{equation} \label{equ:Feedforward}
|\hat{\psi^1}\ldots\hat{\psi^m}\rangle = U_1U_\mathbf{G}U_2U_\mathbf{G}|\psi^1\rangle \ldots U_1U_\mathbf{G}U_2U_\mathbf{G}|\psi^m\rangle
\end{equation}

The $\mathbf{G}$ is Grover's operator~\cite{grover1996fast}, which will flip the amplitude of the quantum state, whose unitary is expressed in Equation~\ref{equ:U_G}. Since the non-linear activation function can not exist on linear quantum systems, the $\mathbf{G}$ holds a different purpose than the activation function, which is to allow VQE to be performed on a different basis. Note that $\mathbf{G}$ is also computationally efficient since it contains only one multi-qubit gate.

\begin{equation} \label{equ:U_G}
U_\mathbf{G} = H^{\otimes m}(2|\psi\rangle\langle \psi|- I)H^{\otimes m}
\end{equation}

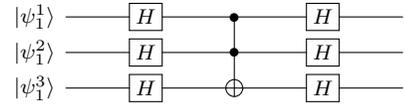
\begin{figure}[htp!]
  \centerline{
    \footnotesize
    \Qcircuit @C=3em @R=.4em {
    & \lstick{|\psi^1_1\rangle} & \gate{H} & \ctrl{1} & \gate{H} & \qw \\
    & \lstick{|\psi^2_1\rangle} & \gate{H} & \ctrl{1} & \gate{H} & \qw \\
    & \lstick{|\psi^3_1\rangle} & \gate{H} & \targ & \gate{H} & \qw
    }
   }
  \caption{Circuit of $\mathbf{G}$}
  \label{fig:g-operator}
\end{figure}

The circuit of $\mathbf{G}$ is shown in Fig.~\ref{fig:g-operator}.

\subsubsection{Measurement}
In the end, quantum measurements on Pauli-Z are performed for every qubit. The measured result is a 1-D tensor where all entries lie in the range of $[-1, 1]$. In order to pass the output into the next block, that is, making the QNet a sequence-to-sequence encoder model, flattened tensors are reshaped to the shape of input.

\subsection{ResQNet}

\begin{figure}[htp!]
  \centering
    \tikzstyle{layer} = [draw, text centered, text width=14em, minimum height=1em, rounded corners]
    \tikzstyle{denseblock} = [layer, fill=violet!10]
    \tikzstyle{outputblock} = [layer, fill=blue!10]
    \tikzstyle{linear} = [layer, fill=orange!10]
    \tikzstyle{qnet} = [layer, fill=green!10]
    \tikzstyle{encode} = [layer, fill=red!20]
    \begin{tikzpicture}
        \node (output) [outputblock] {Output Projection};
        \node (dense) [denseblock] [below of=output] {Linear};
        \node (add) [font=\Large] [below of=dense] {$\oplus$};
        \node (qnet) [qnet] [below of=add] {QNet Encoder};
        \node (linear) [linear] [below of=qnet] {Linear Transformation};
        \node (encode) [encode] [below of=linear] {Embedding Layer};
        
        \begin{scope}[on background layer]
            \node [draw, densely dashed, rounded corners, fit={($(linear)-(3, 0.5)$) ($(add)+(3, 0.4)$)}, label={right:$\times N$}] {};
            \draw[->] (encode) -- (linear);
            \draw[->] (linear) -- (qnet);
            \draw[->] (qnet) -- (add);
            \draw[->] (add) -- (dense);
            \draw[->] (dense) -- (output);
            
            \draw[->] ($(linear)+(0,-0.45)$) -- ($(linear)+(2.7,-0.45)$) -- ($(add)+(2.7,0)$) -- (add);
        \end{scope}
        
    \end{tikzpicture}
  \caption{The ResQNet - the model architecture.}
  \label{fig:resqnet}
\end{figure}
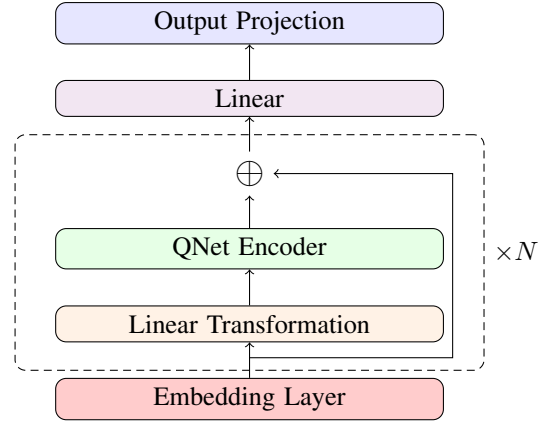

The ResQNet is a quantum-classical hybrid model composed of multiple residual-connected QNet blocks to mimic the architecture of the Transformer Encoder, as shown in Fig.~\ref{fig:resqnet}. In the "Linear Transformation" layer, the input is multiplied by a vector on the dimension of the embedding to scale the weight of each embedding. After QNet Encoder, the previous data are added to the output of QNet and normalized.

The derivation and architecture design are introduced in detail in the following sections.

\subsubsection{Linear Transformation}

The mathematics transformation of this layer is to let classical data be changed into a preferable scale to encode onto the quantum circuit. Practically, the feature matrix is multiplied to a vector of length $d$ to scale variables in the embedding axis, and it can be written as Equation~\ref{equ:resqnet_linear}, where $A$ is the feature matrix and $w$ are the parameters in this linear transformation.

\begin{equation} \label{equ:resqnet_linear}
output = A \times
\begin{bmatrix}
    w_1 & 0 & \dots & 0 \\
    0 & w_2 & 0 & \vdots \\
    \vdots & 0 & \ddots &  \vdots \\
    0 &  \dots & \dots &  w_{d}
\end{bmatrix}
\end{equation}

The design is based on the concept that functionality of QNet can be highly relied on in this model because, instead of the traditional feedforward layer that uses $O(d^2)$ parameters to convert the word embedding to a different meaning, the linear transformation here only uses $O(d)$ parameters with the idea to adjust input for the quantum circuit.

\subsubsection{QNet Encoder}

In our experiment configuration, QNet with a depth of one is used as the QNet block in the ResQNet. It is based on the idea of replicating the architecture of the Transformer Encoder but with the replacement of Multi-head attention blocks and positional feedforward blocks. Our number of quantum measurements is significantly fewer than the previous quantum model, QLSTM, since the transformer architecture is usually not too deep, e.g., the original BERT only used 12 blocks of Transformer Encoder.

\subsection{QNet Complexity Analysis}

In this section, we examine the complexity of QNet and contrast it with BERT and FNet. On a classical computer, the time complexity of a model is an asymptotic function of the number of basic arithmetic operations used. If the task is parallelizable, it would be divided by the number of processes. However, the qubits on quantum computers are often executed independently, so execution time is more likely to equal the depth of the circuit. The circuit depth is determined by observing the critical path, specifically counting the dependent entanglement gates in practice. In addition to the circuit depth, we assess the gate complexity, representing the asymptotic number of basic gates.
Besides execution time, we also analyze the parameter complexity of the model, that is, the asymptotic function of the size of the model.

\begin{table}[htb!]
    \centering
    \caption{Quantum circuit depth analysis of QNet.}
    \begin{tabular}{ l|c|c  }
        \hline
        Circuit block & Gate Complexity & Circuit depth \\
        \hline
        Input Encoding & $O(n \cdot d)$ & $O(1)$ \\
        Mixture Learning &  $O(n^2 \cdot d)$ & $O(n)$ \\
        Positional Feedforward &  $O(n \cdot d)$ & $O(d)$ \\
        \hline
    \end{tabular}
    \label{table:depth}
\end{table}

In Table~\ref{table:depth}, the gate complexity and maximum circuit depth are computed. Here, $n$ is the sequence length, and $d$ is the representation dimension. These complexities are estimated as follows.

\begin{itemize}
  \item The circuit depth is $O(1)$ in the Input Encoding because there are only two rotation gates on every qubit.
  \item According to Theorem~2 in~\cite{892140} and the fact that there is no dependence on each Quantum Fourier Transform in Mixture Learning layer, the depth of Mixture Learning layer is $O(n)$.
  \item In the QNet Positional Feedforward layer, each Grover Operator contains an n-qubit Toffoli gate, and the study~\cite{PhysRevA.87.062318} shows that the Toffoli gate can be decomposed with linear circuit depth, that is, $O(d)$.
\end{itemize}

\begin{table}[htb!]
    \centering
    \caption{Comparison of attention layer time complexity and parameter complexity.}
    \begin{tabular}{ l|c|c  }
        \hline
        Type & Execution Time & Parameter\\
        \hline
        Mixture Learning & $O(n)$ & $O(d)$\\
        Self-Attention &  $O(n^2 \cdot d)$ & $O(d^2)$\\
        FFT & $O(n \cdot  d (\log n + \log d))$ & $O(1)$\\
        \hline
    \end{tabular}
    \label{table:attentions}
\end{table}

In Table~\ref{table:attentions}, QNet Mixture Learning is compared with Self-Attention from Transformer and Fast Fourier Transform from FNet. In Table~\ref{table:feedforward}, the QNet Feedforward Layer is compared with the Positional-wised Feedforward Network from Transformer. The Positional-wised Feedforward Network is used in almost all Transform-based architecture, including BERT and FNet.

\begin{table}[htb!]
    \centering
    \caption{Comparison of feedforward layer time complexity and parameter complexity.}
    \begin{tabular}{ l|c|c  }
        \hline
        Type & Execution Time & Parameter \\
        \hline
        QNet Feedforward & $O(d)$ & $O(d)$\\
        Positional-wised &  $O(n \cdot d^2)$ & $O(d^2)$ \\
        \hline
    \end{tabular}
    \label{table:feedforward}
\end{table}

The model complexity is the addition of attention layer complexity and feedforward layer complexity. Based on Table~\ref{table:attentions} and Table~\ref{table:feedforward}, the overall model operation complexity, parameter complexity, and the actual number of parameters of different models in the configuration of embedding dimension 128 and 2 blocks are shown in Table~\ref{table:overall}. Notice that the complexities of the models in the classical domain can be parallelized by dividing the number of processors.

\begin{table}[htb!]
    \centering
    \caption{Comparison of model complexity and parameters numbers}
    \begin{tabular}{ l|c|c|r  }
        \hline
        Model & Execution Time & Param & Measured Param \\
        \hline
        QNet & $O(n+d)$ & $O(d)$ & 2,304\\
        FNet &  $O(d \cdot  n \log n + n \cdot d^2)$ &  $O(d^2)$ & 67,072 \\
        BERT &  $O(n^2 \cdot d + n \cdot d^2)$ &  $O(d^2)$ & 1,122,048 \\
        \hline
    \end{tabular}
    \label{table:overall}
\end{table}

\section{Experiments}

\subsection{Configurations}

We implemented the QNet with TensorFlow~\cite{abadi2016tensorflow} and TensorFlow Quantum~\cite{broughton2020tensorflow}, a quantum machine learning library.
In the experiments, models are used as the backend for the same task. QNet and ResQNet are compared against FNet and BERT-Tiny~\cite{devlin2018bert} with both models under their original configuration. In addition, we use QLSTM as a comparison target of the quantum-related models.

\subsubsection{Hardware}

GPUs and TPUs are not used in the experiments due to insufficient memory to simulate quantum computers. Instead, our models are trained distributively on Taiwania, a supercomputer cluster in National Center for High-performance Computing. The experiments in this work use four computing nodes, each equipped with two 20-core Intel Xeon Gold 6148 CPUs and 384 GB RAM. The MultiWorkerMirroredStrategy of TensorFlow is used as the distribution strategy for synchronous training on multiple workers to train the model in a cluster.

\subsubsection{Optimizer}

We used the Adam optimizer with $\beta_1 = 0.9$, $\beta_2 = 0.98$ and $\epsilon = 10^{-7}$.

In the distributed training environment, the loss of model is reduced to share across multiple workers. Therefore, we have to adjust the learning rate according to batch size and the number of nodes. The adjusted learning rate is defined as the global learning rate in Equation~\ref{equ:global_lr}. The initial learning rate is $3e-4$ in all experiments.

\begin{equation} \label{equ:global_lr}
global\_lr = initial\_lr \times batch\_size \times num\_nodes
\end{equation}

We varied the learning rate over the course of training according to the Cosine Decay strategy~\cite{loshchilov2017sgdr} in Equation~\ref{equ:lr}. In this work, $\alpha = 10^{-2}$.

\begin{equation} \label{equ:lr}
lr = global\_lr \times ( \frac{1}{2} (1 + cos( \frac{step}{total\_steps}\pi))(1 - \alpha) + \alpha)
\end{equation}

\subsection{Tasks \& Performance Results}

In our experiments, we focus on NLP tasks that have short sequence input. Here, 27\% of data are used for testing while others are used for training. All input sentences will be converted into lowercase and punctuation stripped, and if the sentences in the dataset have not been through tokenization, the consecutive words will be separated by white space. The models are evaluated on text classifications, review score prediction regression, and named entity recognition with input limited to a maximum of 8 tokens. All models are trained with 5 epochs with a batch size of 128, and each epoch has 100 steps.

\begin{table}[htb!]
    \centering
    \caption{Statistics for the text datasets}
    \begin{tabular}{ l|c|c|c|c  }
        \hline
        Dataset & Classes & Size & Length & $|V|$ \\
        \hline
        StackOverflow & 20 & 20,000 & 8.31/34 & 22,956 \\
        ColBERT & 2 & 200,000 & 12.81/22 & 74,010\\
        RentTheRunway & N/A & 192,462 & 7.21/118 & 15,454 \\
        MSRA & 7 & 299,077 & N/A & 4,748\\
        \hline
    \end{tabular}
    \label{table:dataset-statistic}
\end{table}

The statistics of datasets are shown in Table~\ref{table:dataset-statistic}. In the table, Length contains two numbers representing the mean and max length of texts respectively, and $|V|$ is the vocabulary size.

\begin{table*}[htb!]
    \centering
    \caption{The comparison of models when evaluating on ColBERT, StackOverflow, RentTheRunway, and MSRA dataset.}
    \begin{tabular}{ |l|c|c|c|c|c|c|r|r|r|  }
        \hline
        Model & Embeddings & Blocks & ColBERT & StackOverflow & RentTheRunway & MSRA & Qubits & Measurements & Params \\
        \hline
        \textbf{QNet} & 2 & 1 & 89.84 & 16.64 & 1.7765 & 0.0223 & 16 & 1 & 18 \\
        \textbf{ResQNet} & & & \textbf{90.48} & \textbf{30.78} & 1.6910 & \textbf{0.7261} & 16 & 1 & 20 \\
        QLSTM & & & 90.15 & 12.42 & \textbf{1.4293} & 0.5342 & 2 & 48 & 64 \\
        \hline
        \textbf{QNet} & 2 & 2 & 86.89 & 18.20 & 1.7260 & 0.0555 & 16 & 1 & 36 \\
        \textbf{ResQNet} & & & \textbf{91.17} & \textbf{30.08} & 1.7059 & \textbf{0.5318} & 16 & 2 & 40 \\
        QLSTM & & & 90.46 & 10.46 & \textbf{1.4115} & 0.0012 & 2 & 96 & 128 \\
        \hline
        \textbf{QNet} & 3 & 1 & 88.83 & 12.43 & 1.6914 &  0.0783 & 24 & 1 & 27 \\
        \textbf{ResQNet} & & & \textbf{90.99} & \textbf{41.41} & \textbf{1.6622} & \textbf{0.6611} & 24 & 1 & 30 \\
        QLSTM & & & 90.31 & 11.87 & 2.1001 & 0.1032 & 3 & 48 & 120 \\
        \hline
        FNet & 256 & 6 & 84.45 & \textbf{5.69} & 1.7161 & 0.0056 & N/A & N/A & 795,648 \\
        BERT-Tiny & 128 & 2 & \textbf{87.23} & 5.38 & \textbf{1.6714} & \textbf{0.4603} & N/A & N/A & 1,122,048 \\
        \hline
    \end{tabular}
    \label{table:experiment_results}
\end{table*}

The descriptions of the experimented tasks and corresponding datasets are as follows. Note that the number of parameters reported in Table~\ref{table:experiment_results} is the parameters of the whole model without word embeddings since each task has a different $|V|$.

\subsubsection{Text Classifications}

ColBERT~\cite{fw8e-z983-21} dataset is a binary classification task for humor detection. It consists of 100,000 positive and 100,000 negative short formal texts.

Binary Cross Entropy loss is used for ColBERT humor detection, and results are listed in Table~\ref{table:experiment_results}. As can be seen in the table, QNet achieves an accuracy of 89.84\% with an embedding dimension of two and block depth of one, while BERT-Tiny reaches an accuracy of 87.23\%. We observed that in the ColBERT dataset, there are few duplicated words in a single sentence, but there are some tokens often repetitively shown in sentences such as "Why", "your", and "is". Thus, recognizing representative tokens while ignoring massive trash information is the key to success in this task. However, the self-attention mechanism may lose focus on the impactful words seldom shown because these word embeddings have not been pre-trained on a large dataset to get the rough meaning and failed to reach an ideal accuracy.

StackOverflow~\cite{xu-etal-2015-short} dataset is a multi-class classification task to classify questions on StackOverflow.com. It consists of randomly selected 20,000 question titles from 20 different tags.

Categorical Cross Entropy loss is used for StackOverflow question classification, and results are listed in Table~\ref{table:experiment_results}. As can be seen in the table, the QNet and ResQNet outperformed other models. This dataset is more about the detection of the keywords than the relationship of the tokens. The result might be caused by QNet and ResQNet having the benefit of peeping the orienting of the sentences by the entangled quantum state that shares the information across multiple tokens to learn the representation of tokens.

\subsubsection{Review Score Prediction}
In the RentTheRunway~\cite{10.1145/3240323.3240398} dataset, data are the measurements of clothing fit. The review summary is taken as input, and models are asked to predict the rating. The score prediction tasks are usually not classification tasks but regression tasks; the outputs of the model only need to be as close as possible to the labels.

We viewed score prediction tasks as regression tasks, so mean square error is used to optimize the model and as the evaluation metric on test data. The configurations and results of the RentTheRunway review rating prediction are listed in Table~\ref{table:experiment_results}. As observed in the table, that QLSTM has a lower error rate than QNet is reasonable because it uses plenty of VQE circuits in each LSTM gate to make data transform multiple times. Despite that, QNet and ResQNet still have compelling results compared to the state-of-the-art classical models. In this task, context is more important than keywords, so QNet and ResQNet could have the advantage of the high dimension of Hilbert space that spans multiple qubits.

\subsubsection{Named-Entity Recognition}
MSRA~\cite{levow2006third} dataset is a Chinese named-entity recognition dataset with a BIO tagging format~\cite{ramshaw1999text}. In Named-Entity Recognition tasks, most of the tokens will get an O-tag, a tag indicating that the tokens do not belong to any chunk, making the distribution of labels in the dataset heavily imbalanced.

The cross-entropy loss is applied to every token to classify the entity type of the token in the MSRA task, and results are listed in Table~\ref{table:experiment_results}. The F1-score of non-O tags is used to measure the performance. This task is the most complex in this work since the classifier needs to assign tags to every token with low embedding dimensions. Moreover, imbalanced inputs make models often emit O-tag as output which does not produce meaningful results, and such models will have low F1-score in non-O tag as shown in the table.



\subsection{Ablation Study}

To prove the effectiveness of the QNet structure, we use an ablation study under the configuration of the embedding size of 2 and encoder blocks of 2 to test each part independently. Two types of results are demonstrated in Table~\ref{table:ablation_study}: positional feedforward only,  and mixture learning only. 

\begin{table}[htb!]
    \centering
    \caption{The ablation study of QNet.}
    \begin{tabular}{ |l|c|c|c|  }
        \hline
        Configuration & ColBERT & StackOverflow & MSRA \\
        \hline
        Positional Feedforward Only & 87.34 & 15.78 & 0 \\
        \hline
        Mixture Learning only & 83.44 & 14.76 & 0.0857 \\
        \hline
        QNet & 86.89 & 18.20 & 0.0555 \\
        \hline
    \end{tabular}
    \label{table:ablation_study}
\end{table}

It can be seen in Table~\ref{table:ablation_study} that without the Mixture Learning layer, Positional Feedforward performs poorly on high relational input tasks. For example, it can not learn anything from the MSRA dataset. And without Positional Feedforward, Mixture Learning showed a shortage of keyword-critical tasks such as ColBERT.

\subsection{Noise Tolerance}

To exhibit how QNet is resistant to the noises in NISQ-era devices, we use the depolarize function from the Cirq library to observe the change of training loss in different amounts of noise. The experiment uses the ColBERT dataset, and QNet is trained for one epoch and batch size of 8.

\begin{figure}[htbp]
\centerline{
    \includegraphics[width=7cm]{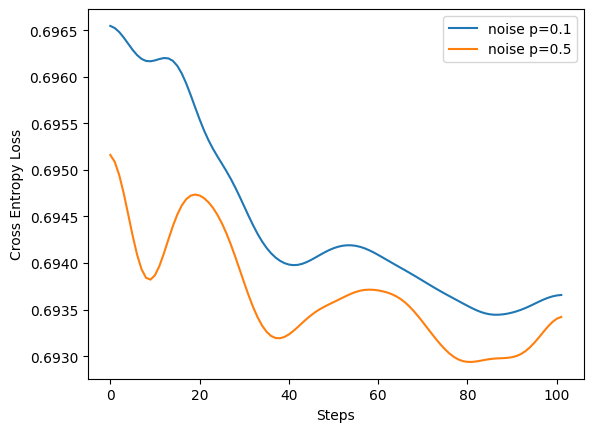}
}
\caption{The training loss of noisy circuit}
\label{img:noise}
\end{figure}

In Fig.~\ref{img:noise}, the $p$ in the legend is the probability of error generated by the depolarize function. We can see that the training curve of $p=0.5$ is much more jitter than training $p=0.1$, but the loss is descending and reaches about the same value in the end.

\section{Conclusion}

This work presents the QNet, the quantum-native sequence encoder model. QNet features short circuit depth and low qubit usage, making it feasible in near-term quantum computers. In addition, we introduce the ResQNet, the residual connected QNet, to use it in the current stage of quantum computing resources. For various NLP tasks, including text classifications, review score prediction regression, and named-entity recognition, the QNet and ResQNet have advantages compared to classical machine learning models with large embedding dimensions.

We look forward to revealing how quantum entanglement relates to NLP. We want to analyze and explain the result of neural networks to know how high dimension Hilbert space spanned by entanglement gates in QNet blocks assists the model in holding sparse information since, by the mathematics formula, QNet blocks can potentially contain the information of relation among many tokens. However, it is challenging and requires strenuous efforts, especially in hybrid models concerning two different device types and the fact that there are few proposed methods to investigate quantum machine learning models. Also, we are excited about training QNet in the large-scale NISQ quantum computer in the future. When the scale of quantum computers is ready to employ residual connections, we plan to extend the ResQNet to a pure quantum model as QNet. 

\printbibliography
\end{document}